\documentclass[10pt,letterpaper]{article}

\usepackage{cogsci}
\usepackage{amsmath}
\usepackage{hyperref}
\cogscifinalcopy 

\usepackage{pslatex}
\usepackage{apacite}
\usepackage{float} 

\usepackage{graphicx}
\usepackage{colortbl}
\usepackage{booktabs}
\usepackage[table]{xcolor}
\usepackage{siunitx} 

\usepackage{tabularx}
\usepackage{graphicx}

\usepackage{xcolor}

\newcommand\blfootnote[1]{
  \begingroup
  \renewcommand\thefootnote{}\footnote{#1}
  \addtocounter{footnote}{-1}
  \endgroup
}

\title{Can (A)I Change Your Mind?}

\author{
\begin{tabular}{cccccc}
\multicolumn{3}{c}{Miriam Havin$^{* 1,2}$} & \multicolumn{3}{c}{Timna Wharton Kleinman$^{* 3}$} \\
\multicolumn{3}{c}{\href{mailto:miriam.havin@mail.huji.ac.il}{miriam.havin@mail.huji.ac.il}} & \multicolumn{3}{c}{\href{mailto:timna.kleinman@mail.huji.ac.il}{timna.kleinman@mail.huji.ac.il}} \\
\\
\noalign{\vskip -7pt}
\multicolumn{2}{c}{Moran Koren$^{4}$} & \multicolumn{2}{c}{Yaniv Dover$^{2,5}$} & \multicolumn{2}{c}{Ariel Goldstein$^{1,5}$}\\
\multicolumn{2}{c}{\href{mailto:korenmor@bgu.ac.il}{korenmor@bgu.ac.il}} & \multicolumn{2}{c}{\href{mailto:yaniv.dover@mail.huji.ac.il}{yaniv.dover@mail.huji.ac.il}} & \multicolumn{2}{c}{\href{mailto:ariel.y.goldstein@mail.huji.ac.il}{ariel.y.goldstein@mail.huji.ac.il}} \\
\end{tabular}\\
\\
\noalign{\vskip -8pt}
\small $^1$Department of Cognitive and Brain Sciences, The Hebrew University of Jerusalem, Israel\\
\noalign{\vskip -2.5pt}
\small $^2$The Federmann Center for the Study of Rationality, The Hebrew University of Jerusalem, Israel\\
\noalign{\vskip -2.5pt}
\small $^3$School of Computer Science and Engineering, The Hebrew University of Jerusalem, Israel\\
\noalign{\vskip -2.5pt}
\small $^4$Industrial Engineering and Management Department, Ben Gurion University, Israel\\
\noalign{\vskip -2.5pt}
\small $^5$The Hebrew University Business School, Jerusalem, Israel
}

\begin{document}

\maketitle

\begin{abstract}
The increasing integration of large language models (LLMs) based conversational agents into everyday life raises critical cognitive and social questions about their potential to influence human opinions.
Although previous studies have shown that LLM-based agents can generate persuasive content, these typically involve controlled English-language settings. Addressing this, our preregistered study explored LLMs' persuasive capabilities in more ecological, unconstrained scenarios, examining both static (written paragraphs) and dynamic (conversations via Telegram) interaction types. Conducted entirely in Hebrew with 200 participants, the study assessed the persuasive effects of both LLM and human interlocutors on controversial civil policy topics. Results indicated that participants adopted LLM and human perspectives similarly, with significant opinion changes evident across all conditions, regardless of interlocutor type or interaction mode. Confidence levels increased significantly in most scenarios. These findings demonstrate LLM-based agents' robust persuasive capabilities across diverse sources and settings, highlighting their potential impact on shaping public opinions.\blfootnote{$^{*}$These authors contributed equally to this work.}

\end{abstract}

\section{Introduction}
Conversational large language models (LLMs) have rapidly evolved from a novel technology to an integral part of daily human interaction. As LLM-driven conversational agents (such as ChatGPT, Claude, and Gemini \cite{chatgpt,claude,gemini}) become increasingly embedded in consumer technology, workplace tools, healthcare services, and educational platforms, their ability to persuade human users raises important cognitive and social questions \cite{rogiers_persuasion_2024}. While recent studies have demonstrated that LLMs can generate persuasive content (e.g., \citeA{breum_persuasive_2024,costello_durably_2024}), the extent to which their persuasive capabilities compare to those of humans, particularly in real-world, unconstrained discussions, remains underexplored.

Importantly, the question of machine persuasion is not new. Decades of research in human-computer interaction (HCI) and communication provide a relevant foundation. These studies demonstrate that people often respond to computers and digital agents in fundamentally social ways, at times even applying social norms to them \cite{fong2003survey, nass2000machines, nass1994computers, reeves1996media}. Moreover, real-world persuasion often unfolds dynamically through open-ended conversations, where interactivity and engagement shape outcomes. Theories of dialogue and communicative alignment, such as the interactive alignment model \cite{pickering2004toward} and Communication Accommodation Theory \cite{giles1991accommodation}, highlight how interlocutors adapt to one another in real time, converging or diverging in language, style, and stance. These insights suggest that an agent's adaptability, not just the content of its arguments, plays a critical role in shaping user receptivity, particularly in dynamic interactions.

Despite these insights, much of the recent literature on LLM-based persuasion has focused on static, one-shot persuasive messages \cite{bai_artificial_2023,karinshak_working_2023} or structured, controlled interactions \cite{costello_durably_2024}, with limited attention given to the dynamic, interactive nature of real-world dialogue.

To address this gap, we conducted a series of experiments designed to systematically compare the persuasive effectiveness of human and LLM-driven interactions in both static (i.e.,  one-time, non-interactive message exposure) and dynamic (i.e., real-time, back-and-forth conversation) settings. Our primary goal was to assess whether interacting with an agent in an open, naturalistic environment could lead to opinion change, and how this compares to persuasion by humans. By "naturalistic environment", we refer to conditions that mirror real-world interactions: Conversations are open-ended, without constraints on message length or frequency, reflecting the fluid nature of everyday discourse. Participants and LLMs are allowed to send multiple messages consecutively without waiting for replies, capturing the organic flow of conversations where thoughts are often shared in quick succession. Additionally, all interactions took place on Telegram, a familiar and widely used messaging platform, further grounding our research in the context of everyday communication tools. These aspects ensure our experimental setup closely replicates the unpredictable and dynamic nature of genuine human dialogue, thereby enhancing the ecological validity of our research.

Our experiment assesses changes in opinion and confidence on controversial civil policy questions relevant to daily life, such as "Do you think there should be a tax on food products with a high amount of saturated fat?". The topics we use are both controversial and relatable, making them ideal for studying how persuasion unfolds in natural discussions. Additionally, our study is conducted entirely in Hebrew, broadening the scope of LLM persuasion research beyond English-dominant studies.

Our study consisted of three key experimental phases, each measuring changes in opinion and confidence in that opinion. First, we validated our experimental framework and created a benchmark, by testing human-human interactions in a dynamic setting, ensuring that the infrastructure functioned as intended and that opinion shifts could be reliably measured. Second, we examined whether LLM-driven persuasion could also lead to opinion change in an unconstrained, real-world setting. Finally, we directly compared the persuasive effectiveness of LLM and human interlocutors under these ecological conditions, using human persuasion as a benchmark. Additionally, we explored a novel and important question: does interaction style—dynamic dialogue versus static, one-time messaging—affect the persuasive impact of LLM-based agents differently than it does for humans? While prior research has identified factors that contribute to LLM persuasiveness, such as personalization and interactivity \cite{matz_potential_2024,salvi_conversational_2024}, this direct comparison of dynamic and static LLM interactions remains largely unexamined \cite{jones_lies_2024}. 

Our findings contribute to the growing body of research on LLM-human persuasion by addressing two key questions: (1) Can an LLM be persuasive in an ecological, non-constraint, natural conversation?
(2) How does it compare with other persuasion contexts- human-human ecological interactions and LLM-generated messages presented without interaction?
By systematically examining these questions, we provide a nuanced view of how interaction mode (static vs. dynamic communication) and dyad type (human-human vs. human-bot pairings) affect persuasion, offering practical and theoretical insights into the evolving role of LLM-based agents in shaping human opinions on everyday issues.

\section{Methods}
\subsection{Overview}
Our study consists of three separate experiments as well as preparation and data collection phases, 
and will be described in this chapter. All experiments in this study were pre-registered (see pre-registration \href{https://aspredicted.org/gnfg-46py.pdf}{here}\footnote{\href{https://aspredicted.org/gnfg-46py.pdf}{aspredicted.org/gnfg-46py.pdf}}).

\subsection{Question Selection}
The selection process for the experiment's questions was aimed at identifying controversial topics within Israeli society. Initially, a list of 20 controversial questions formatted for 'yes' or 'no' responses was compiled. To refine the selection, 46 participants from the Hebrew University's online experiment participation platform provided answers and rated their confidence on a scale of 1 to 10 for each question. 

We chose the 5 most divisive questions with the highest confidence ratings to ensure they were both contentious and significant to the participants, making them suitable for studying the impact of discussion on opinion and confidence levels. The final five questions were: (1) In your opinion, should companies that encourage working from home be given grants to reduce environmental impact? (2) In your opinion, should child allowances be increased? (3) In your opinion, does the court in Israel currently have an imbalanced amount of power in relation to the executive branch? (4) In your opinion, should a tax be imposed on food products with high saturated fat content? (5) In your opinion, can a building be required to undergo urban renewal (TAMA) if some apartment owners oppose the construction?

\subsection{Experiments}
The study was structured into two preliminary experiments to validate the framework. Experiment 1 involved a human-human dynamic in which 40 participants (20 dyads) engaged in discussions on five topics via the Telegram platform to establish a baseline for participant interaction in a dynamic setting. Experiment 2 explored the human-bot dynamic, involving 28 participants who interacted with a GPT-4 \cite{openai2023gpt} based bot under similar conditions (full prompt can be seen \href{https://github.com/Ariel-Goldstein-Lab/Can-AI-Change-Your-Mind}{here}\footnote{\href{https://github.com/Ariel-Goldstein-Lab/Can-AI-Change-Your-Mind}{github.com/Ariel-Goldstein-Lab/Can-AI-Change-Your-Mind}}). The goal here was to assess whether LLM-driven persuasion could also effect opinion changes in a dynamic setting. 

Following Experiment 1 \& 2, Experiment 3 employed a 2×2 factorial design examining the effects of dyad type and interaction mode. The study included 200 participants (F = 114), ranging in age from 18 to 50 years (mean = 29.17, SD = 7.13). Participants were recruited through online public groups and randomly assigned to one of four experimental conditions. The first factor, dyad type, compared human-human interactions with human-bot interactions. The second factor, interaction mode, contrasted static interactions, where participants just read paragraphs pre-written by their conversation partner, with dynamic interactions, conducted in real time through Telegram discussions. Each of these four conditions - human-human static, human-human dynamic, human-bot static, and human-bot dynamic - was tested with 50 participants. To ensure clarity and relevance for the participants, all materials, instructions, and interactions were provided in Hebrew. All participants were informed in advance whether they would be interacting with a human or a bot partner in
order to avoid contamination of results due to participants detecting
the bot mid-conversation and disengaging.

\begin{figure}
    \begin{center}
    \includegraphics[width=1\linewidth]{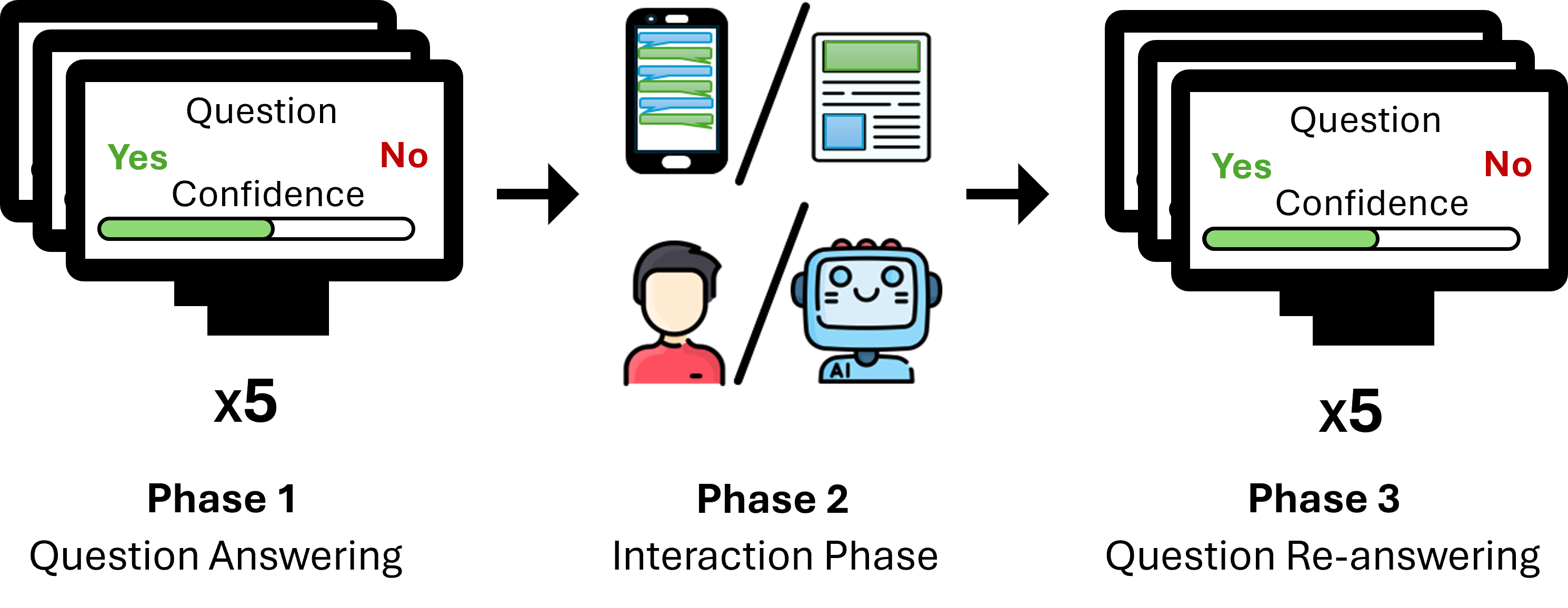}
    \end{center}
    \caption{Experiment Overview} 
    \label{experiment_overview}
\end{figure}

The procedure for each condition in both the preliminary experiments (Exp. 1 \& 2) and the main experiment (Exp. 3) followed a consistent three-phase structure  (see Fig. \ref{experiment_overview}). (1) Participants initially answered the five chosen questions and rated their confidence (on a scale of 1-10, full questionnaire in Hebrew can be seen  \href{https://jatos.mindprobe.eu/publix/io4NLIkFZbs}{here}\footnote{\href{https://jatos.mindprobe.eu/publix/io4NLIkFZbs}{jatos.mindprobe.eu/publix/io4NLIkFZbs}}). To minimize ordering effects, the presentation of these questions was randomized. (2) The participants participated in an interaction - either talking to their partner on Telegram (dynamic) or reading a pre-written paragraph by their partner (static). (3) After the discussion or reading phase, participants re-answered the original questions and reassessed their confidence levels to measure any changes resulting from the interaction.

More specifically, in the dynamic conditions, participants engaged in discussions about the five topics via a Telegram group with their randomly assigned partner. They were given no instructions other than to ensure that all topics were discussed and to conclude the conversation once they felt all issues had been adequately addressed. In the static condition, instead of engaging in discussions, each participant received a unique set of five paragraphs, one corresponding to each policy question. Every set of paragraphs was created by a different author - either a specific human writer or an LLM with a specific prompt - ensuring that each participant encountered a distinct perspective. To maintain internal consistency within each set, all five paragraphs given to a particular participant were authored by the same source (either a single human writer or an AI model maintaining context throughout a single conversation). This approach ensured that while different participants encountered different viewpoints and arguments, each participant received a cohesive set of responses reflecting consistent reasoning and style across all five questions. See more about static collection in the next sub-chapters.
\subsection{Human Static Collection}
For the human static condition, 50 participants were recruited prior to the start of the experiment to respond to the same five questions, first by providing a binary yes/no answer and a confidence rating, followed by writing 3-4 sentence opinions on each question. Participants who did not adhere to these guidelines were excluded from further analysis.

\subsection{Dynamic and Static Bot Structure}
\begin{figure}
    \begin{center}
    \includegraphics[width=1\linewidth]{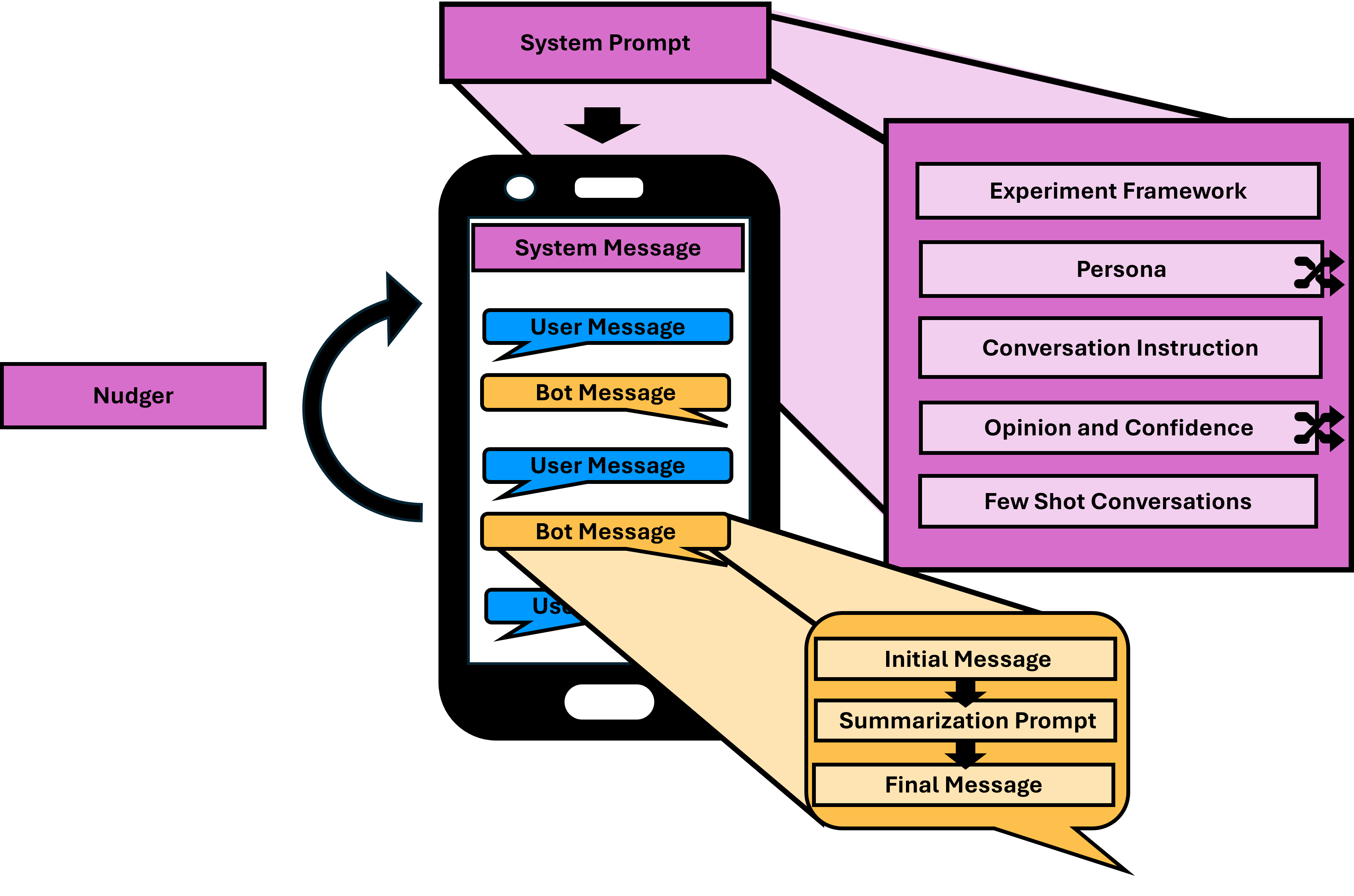}
    \end{center}
    \caption{Dynamic Bot Overview} 
    \label{bot overview}
\end{figure}

For the dynamic bot, we utilized a GPT-4 powered system to facilitate an unstructured conversational dynamic, entirely in Hebrew (Fig. \ref{bot overview}). The initial system prompt was structured into five parts: an introduction to the experiment framework, a persona for the bot (including name, gender, and occupation), instructions for conducting the conversation (be assertive, use examples, acknowledge counterarguments,
use concise everyday language etc.), a list of discussion questions with initial opinions and confidence levels, and two example conversations. The persona, opinions, and confidence were randomized for each participant. 

As users entered the conversation group, they received an automatic system message with conversation instructions and a list of the five questions. The conversation flow was uncontrolled, allowing for natural interaction dynamics.
Each bot message underwent a double iteration process with the GPT-4o model. The initial message composed by GPT-4 was sent to a GPT-4o model, which then summarized and rephrased the message in a manner that mirrored the user’s communication style. This ensured that the responses were both concise and stylistically aligned with the participants' inputs. Additionally, every 5 minutes, an automatic script checked for user activity; if no messages were sent during that period, a reminder was sent to ensure continued engagement.

The static bot configuration was designed to mimic the dynamic bot's parameters as closely as possible but was limited to creating a single paragraph response per question. This paragraph was generated using a similar prompt that included the experiment framework, a randomized bot persona, and initial opinions and confidence levels for the questions. This approach maintained consistency in bot behavior across different interaction modes.

\section{Results}

Experiments 1 \& 2 were designed to assess whether interactions within our setup could significantly influence participants' confidence and opinions in both human-human and human-bot dynamics. The analysis revealed that dynamic interactions significantly influenced both participants' opinions and confidence, regardless of whether they interacted with humans or bots. 

To reach these conclusions, we conducted two types of analyses (see table \ref{table exp 1-2}). First, we examined opinion changes by calculating the proportion of changed responses per participant and assessing the overall proportion of changes within each condition, using confidence intervals to determine statistical significance. Second, we investigated confidence changes through paired t-tests, comparing pre- and post-interaction measurements. The confidence intervals excluding zero for opinion changes and the significant t-test results for confidence demonstrate the robust impact of dynamic interactions on both measures.

\definecolor{botcolor}{RGB}{65,105,225}  
\definecolor{humancolor}{RGB}{255,99,71}  

\begin{table}[h]
\small
\centering
\caption{Changes in Opinions and Confidence}
\label{table exp 1-2}
\resizebox{0.9\linewidth}{!}{
\begin{tabular}{l@{\hskip 3pt} p{1.8cm}@{\hskip 3pt} p{1.6cm}@{\hskip 3pt} p{1.5cm}@{\hskip 3pt} p{1.8cm}@{\hskip 3pt} p{1.3cm}}
\toprule
& \multicolumn{2}{c}{Opinion Change} & \multicolumn{3}{c}{Confidence t-test} \\
\cmidrule(r){2-3} \cmidrule(l){4-6} 
Condition & Changed (\%) & CI & $\Delta$Mean & t(df) & p \\
\midrule
\rowcolor{botcolor!15}
Bot & 23.6 & [2,44] & 0.89 & 3.61(149) & $<$.001 \\
\rowcolor{humancolor!15}
Human & 21 & [4,38] & 0.62 & 2.94(199) & $<$.01 \\
\bottomrule\\
\multicolumn{6}{l}{\small Note: CI = 99\% Confidence Interval. $\Delta$Mean = change in confidence.}
\end{tabular}
}
\end{table}

In Experiment 3, we analyzed opinion and confidence changes as well, assessing the significance of each condition separately. Since conditions were randomly assigned, we also examined interaction effects between them. Additionally, we investigated whether participants were more likely to revise their answers after initial disagreement versus agreement.

\begin{figure}[t]
    \begin{center}
    \includegraphics[width=1\linewidth]{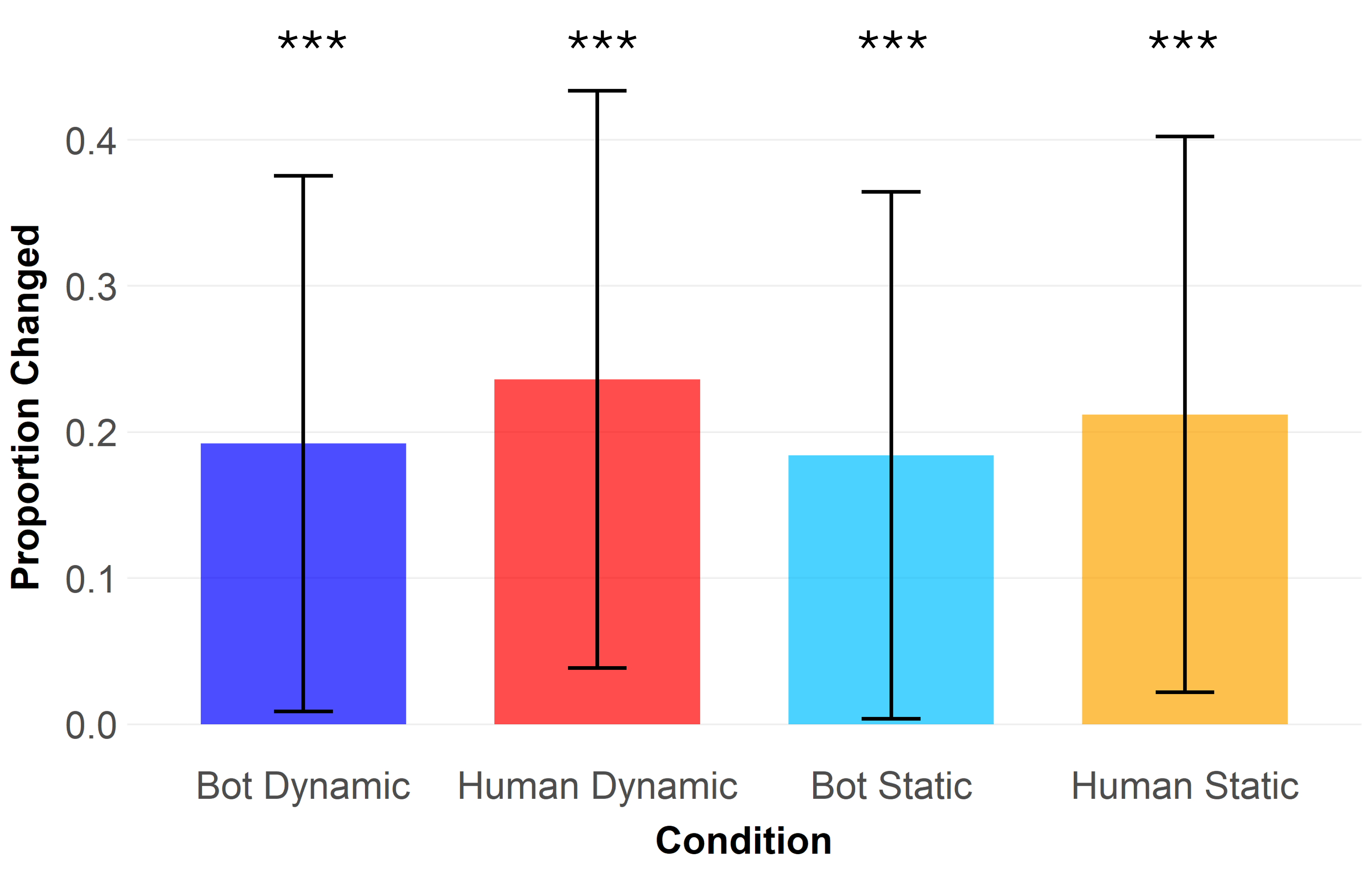}
    \end{center}
    \caption{Proportion of changed opinions across experiments. Error bars represent 99.9\% confidence intervals}
    \label{opinion change}
\end{figure}

For all four conditions, we found that participants changed opinions on a significant portion of the questions (see Fig. \ref{opinion change}). Specifically, for the Human-Bot Dynamic condition, the proportion of changed responses was 19.2\%, 99.9\% CI $[0.8, 37.5]$. For the Human-Human Dynamic condition, the proportion was 23.6\%, 99.9\% CI $[3.8, 43.3]$. For the Human-Bot Static condition, the proportion of changed responses was 18.4\%, 99.9\% CI $[0.3, 36.4]$. For the Human-Human Static condition, the proportion was 21.2\%, 99.9\% CI $[2.1, 40.2]$.
These findings confirm that participants in each condition significantly reconsidered and modified their initial opinions following the interactions, with the confidence intervals clearly indicating the statistical significance of these changes.

In the next stage, we analyzed the differences in opinion change between the four conditions. Our results consistently showed that neither dyad type (human-human vs. human-bot) nor interaction mode (dynamic vs. static) played a significant role in shaping opinion change. Meaning interacting with a bot was as meaningful as interacting with another human, and reading a paragraph was as meaningful as talking to a partner. Frequentist t-tests found no significant differences between conditions, and effect size calculations (Cohen’s d) confirmed that any observed differences were negligible.  To further strengthen support for the null hypothesis, Bayesian t-tests provided moderate to strong evidence for similarity, with ($\text{BF}{10} = 3.42$) for dynamic conditions, ($\text{BF}{10} = 7.69$) for static conditions, ($\text{BF}{10} = 9.39$) for human-human dyads, and ($\text{BF}{10} = 13.41$) for human-bot dyads. These results indicate strong support for the null hypothesis, reinforcing that interaction format and dyad type do not meaningfully influence opinion shifts.

\begin{figure}[t]
    \begin{center}
    \includegraphics[width=1\linewidth]{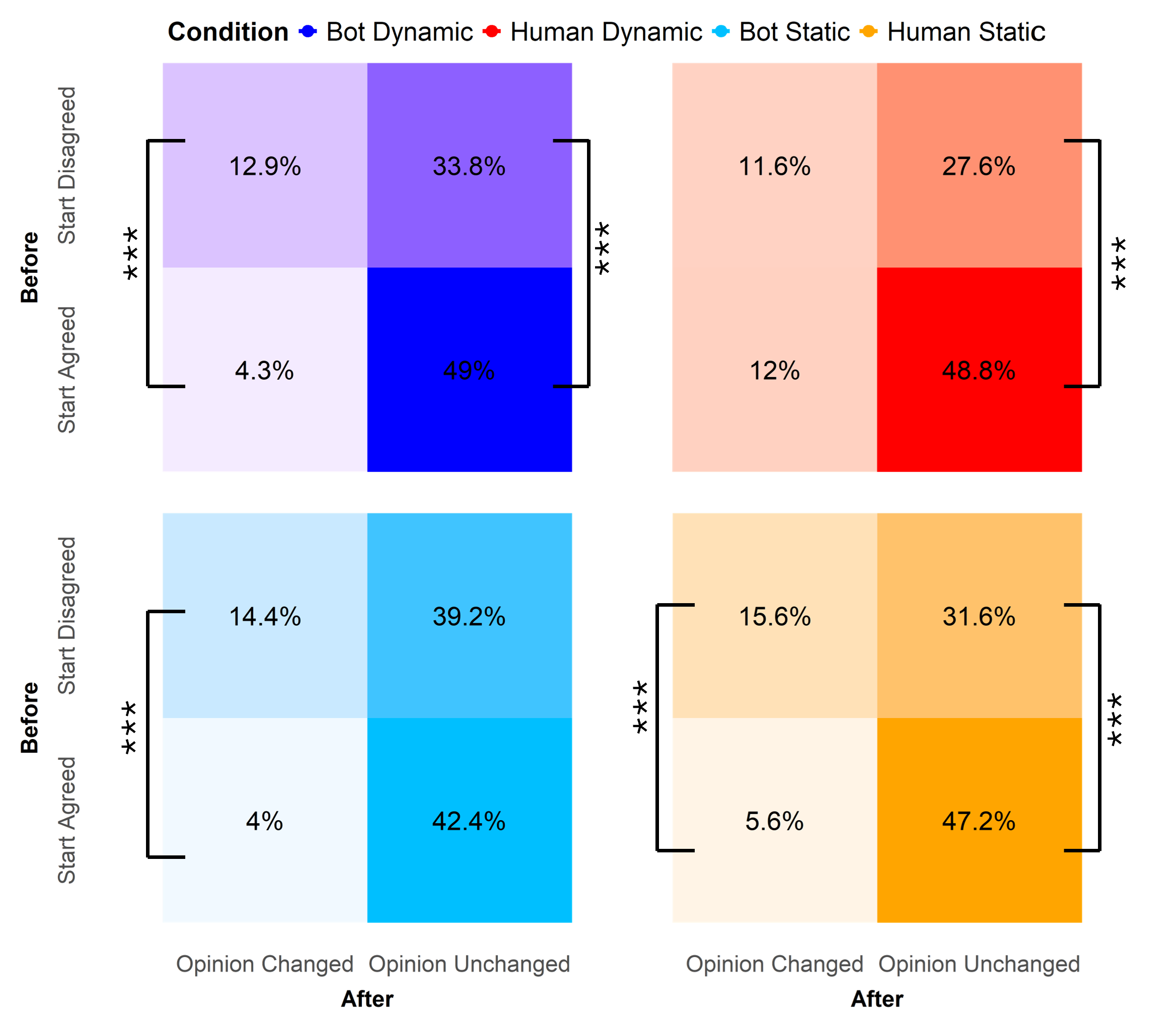}
    \end{center}
    \caption{Proportion of opinion changes by initial agreement with conversation partner across conditions. Each quadrant shows the percentage of participants who changed or maintained their opinions, separated by whether they initially agreed or disagreed with their conversation partner.}
    \label{persuasion matrix}
\end{figure}

Additionally, we examined two aspects of participants opinion change across various interaction settings: 1) whether participants were more likely to change their answers following initial disagreement compared to initial agreement, and 2) whether they were more inclined to maintain their answers when there was initial agreement rather than initial disagreement (as illustrated in Fig. \ref{persuasion matrix}). Our findings indicate that initial disagreements frequently prompted participants to change their responses, a trend particularly pronounced in dynamic settings and bot-mediated interactions. Conversely, initial agreements were more often associated with maintaining responses, especially in dynamic contexts. The analysis indicates that opinion changes were driven by conversational dynamics rather than the time interval between questionnaires, demonstrating the impact of interaction patterns on participant persuasion. However, exceptions were observed, such as in the Bot Static condition, where initial agreement did not consistently result in unchanged responses. This suggests that the effects of initial agreement or disagreement on response changes can vary significantly based on the specific dynamics of the interaction or the characteristics of bot involvement.

\begin{figure}[t]
    \begin{center}
    \includegraphics[width=1\linewidth]{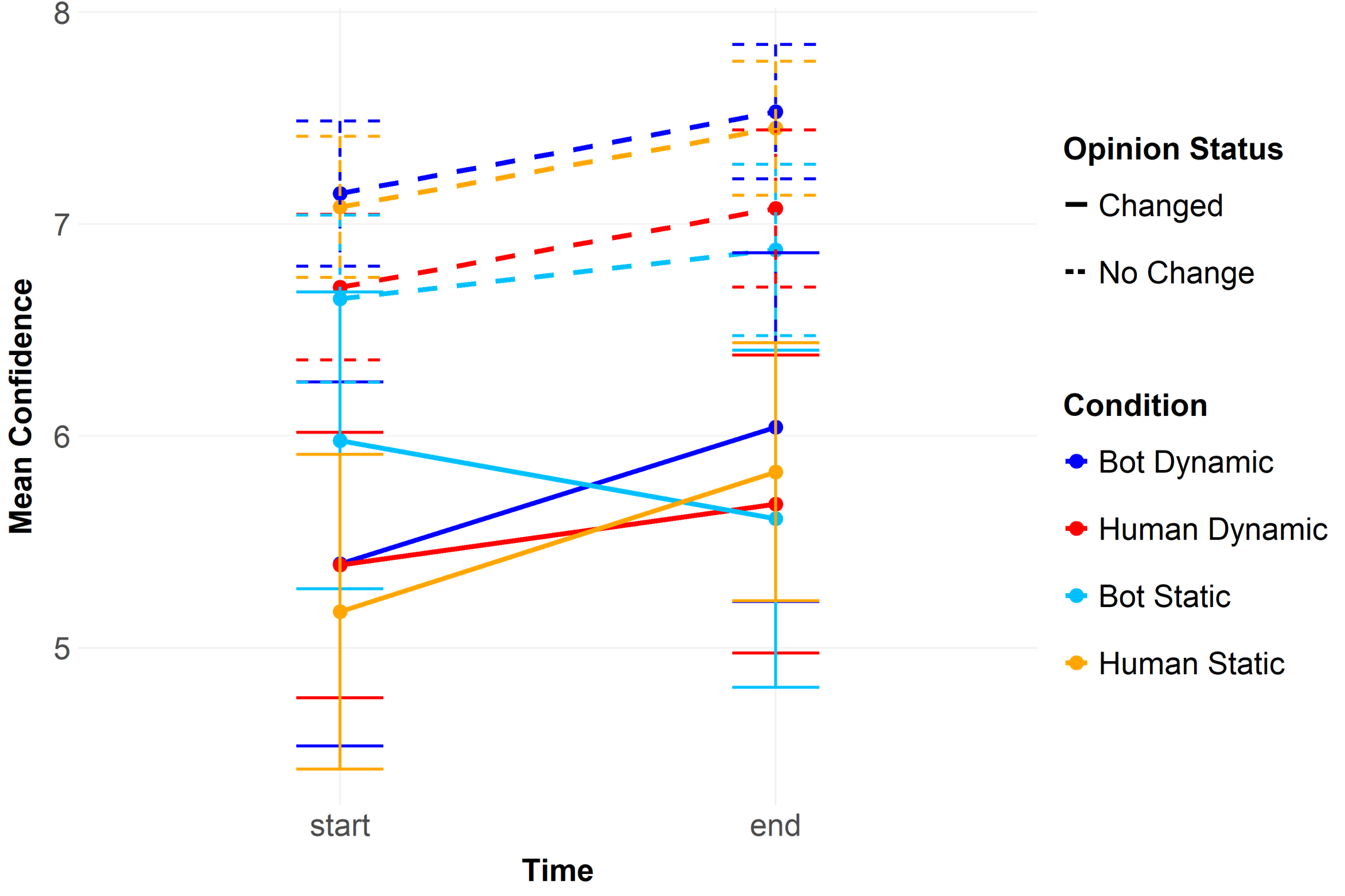}
    \end{center}
    \caption{Mean confidence ratings at the start and end of interactions, shown separately for participants who changed or maintained their opinions across four experimental conditions (Human/Bot × Dynamic/Static)}
    \label{confidence change}
\end{figure}

Next, we analyzed the change in confidence between conditions (see Fig. \ref{confidence change}). We found that confidence increased significantly in all conditions, except for the static bot condition, in which confidence increased marginally. For the Human-Bot Dynamic condition, confidence increased significantly, $t(249) = 2.59, p < 0.01$. For the Human-Bot Static condition, the confidence did not significantly increase, $t(249) = 0.73, p = 0.465$. For the Human-Human Dynamic condition, there was a significant increase in confidence, $t(249) = 2.14, p < 0.05$. For the Human-Human Static condition, confidence also increased significantly, $t(249) = 2.73, p < 0 .01$. 

Additionally, we conducted a mixed-design ANOVA to examine how confidence levels changed over time before vs. after and whether participants changed their opinions. The analysis revealed a significant main effect of opinion change, $F(1, 998) = 67.80$, $p < .001$, $\eta^2_G = .048$, indicating that participants who changed their opinions experienced significantly different confidence levels compared to those who did not. There was also a significant main effect of time, $F(1, 998) = 10.60$, $p = .001$, $\eta^2_G = .003$, suggesting that, overall, participants’ confidence levels significantly increased from before to after the interaction.

However, the interaction between time and opinion change was not significant, $F(1, 998) = 0.01$, $p = .928$, $\eta^2_G < .001$. This indicates that the increase in confidence over time did not differ significantly between participants who changed their opinions and those who did not. In other words, while both time and opinion change individually affected confidence, their combined influence was not significant.

To examine patterns of confidence change, we compared two linear mixed-effects models - with and without interaction between time and condition type. Both models included a random intercept for conversation. When comparing interaction types (static vs. dynamic), BIC values favored the simpler model without interaction terms for both human-human ($\Delta\text{BIC} = 6.83$) and human-bot ($\Delta\text{BIC} = 5.73$) dyads. Similarly, when comparing dyad types, BIC values also supported the simpler models in both dynamic ($\Delta\text{BIC} = 6.89$) and static ($\Delta\text{BIC} = 5.79$) conditions. Across all comparisons, the consistent preference for models without interaction terms (as indicated by lower BIC values) suggests that while there were main effects of time and condition type on confidence ratings, the patterns of confidence change over time remained relatively stable regardless of dyad type or interaction format.

\section{Discussion}
First and foremost, our study demonstrates that in both static and dynamic settings, AI agents can be persuasive, effectively shifting opinions. Crucially, our results generalize prior findings on AI-driven persuasion by showing that these effects hold even in an ecologically valid, real-world conversational setting. Unlike many prior studies that focused on structured, English-language interactions, our study was conducted entirely in Hebrew, emphasizing the robustness of AI's persuasive capabilities across linguistic and cultural contexts \cite{naous_having_2024,ramezani_knowledge_2023,schimmelpfennig_moderating_2024}.

Furthermore, our findings underscore the robustness of LLM persuasion even when participants are fully aware that they are interacting with a bot. While previous research suggests source awareness can reduce LLM credibility \cite{teigen_persuasiveness_2024}, our study shows that persuasion remains effective despite this awareness. This suggests that the quality of the arguments, rather than the source, may be the primary driver of the change in opinion.

Our study also measured two distinct aspects of persuasion: opinion change and confidence shifts. While opinion shifts were evident across conditions, confidence increased significantly in all but the static bot condition. This indicates that conversational engagement, particularly in dynamic settings, reinforces participants’ certainty in their positions, whether or not their opinions change. Future research should further explore the interplay between confidence and persuasion, as increased confidence does not necessarily correlate with greater persuasion success.

A key insight from our findings is that persuasion occurred consistently across all conditions and was unaffected by dyad type or interaction mode. Whether the interaction was with a human or a bot, static or dynamic, the overall patterns of opinion change and confidence shifts remained similar. This challenges assumptions that dynamic engagement is inherently more persuasive than static message exposure.  One possible explanation is that persuasion relies more on the clarity, coherence, and perceived relevance of the arguments than on the mode of delivery. A well-structured static message may contain all the necessary persuasive elements upfront, making additional interaction redundant.

From a real-world perspective, our findings have important implications for understanding how AI-driven persuasion operates in different digital environments. The static condition mirrors scenarios like social media posts, where individuals encounter brief persuasive messages without interaction. The dynamic condition is more akin to real-time conversations in messaging apps or forums. Our results suggest that LLMs can be persuasive in both formats, reinforcing concerns about their potential influence in online discourse, particularly in politically or socially charged discussions. 

Additionally, preliminary examination shows that static interactions required significantly less time than dynamic interactions, as participants only needed to read a short paragraph rather than engage in an extended discussion. This has important practical implications: if static persuasion is as effective as dynamic persuasion, then shorter, one-time messages could be a more efficient and scalable means of influencing opinions in real-world applications.

Finally, while our study focuses on the effectiveness of LLM persuasion, it also raises ethical concerns. The ability of LLMs to influence opinions so effectively necessitates discussions on responsible AI deployment. Issues such as transparency, manipulation, and misinformation warrant careful consideration, particularly as AI systems become increasingly integrated into everyday communication.

In sum, our study provides strong evidence that AI-driven persuasion is both effective and generalizable across different cultural and linguistic contexts. The fact that persuasion remains consistent across conditions, even with minimal engagement in static settings, highlights the broad applicability of these findings. As AI systems continue to evolve, understanding their persuasive impact will be critical for both policymakers and researchers alike.

\section{Limitations and Future Work}
While our study provides valuable insights into AI-driven persuasion, it also highlights areas for further exploration.

One consideration is the ecological nature of our setup. While allowing natural conversations increased realism it also introduced variability in argument depth across conditions—and in the dynamic setting, variation in the number of arguments as well. Since interactions were not strictly controlled, differences in engagement levels may have influenced the observed effects. Future research should disentangle the key factors that shape meaningful interactions by conducting more controlled experiments that systematically vary argument quantity, interaction duration, and conversational depth.

Another important factor is participants’ awareness regarding who they were interacting with, which, as prior studies suggest, may have influenced their responses \cite{aydin_dissociated_2024, lim_effect_2024, teigen_persuasiveness_2024}. 
Given that all conditions in our study yielded similar persuasion outcomes, this raises the question of whether awareness of interacting with a bot limits LLM-based agents' persuasion effectiveness. If participants had not been explicitly informed that they were engaging with an AI, it is possible that the persuasive impact of the bot could have been even stronger. Beyond credibility effects, further research should explore how LLM awareness influences engagement, resistance to persuasion, and reliance on heuristics. Key factors to examine include one's previous AI experience \cite{lim_effect_2024}, trust in automation, and topic sensitivity.

Our study also focused on immediate persuasive effects without assessing long-term opinion stability. We did not conduct follow-up evaluations (e.g., after 10 days or 2 months) to determine whether observed changes in opinion persisted over time. 
Moreover, while most research on LLM persuasion—including ours—examines belief change within experimental settings, further work is needed to understand how these effects translate into real-world behaviors, where individuals face actual costs and responsibilities.

Future work should move beyond outcome measures and undertake a fine-grained content analysis of the conversation transcripts. By coding each message for argumentative structure, rhetorical strategies, and evidential strength, researchers could uncover what constitutes high-quality argumentation in the different tested conditions. Such analysis would clarify whether similar persuasive outcomes stem from shared argumentative patterns or qualitatively distinct strategies and link the different strategies to established models of argumentation and persuasion \cite{petty1986elaboration, Toulmin_2003, van2004systematic, walton2008argumentation}.

An additional open question is how the bot’s opinion and confidence evolved throughout the discussion. 
Unlike humans, bots do not have genuine opinions or independent reasoning processes and are known to adjust their responses to align with user expectations, a phenomenon known as "sycophancy" \cite{perez_discovering_2022, wei_simple_2024}. Understanding how this tendency influences persuasion effectiveness and user trust remains an important open question. Future work should investigate whether dynamic LLM responses enhance persuasion by fostering engagement or undermine trust by appearing inconsistent.

Finally, while we used a specific prompting approach and model size, prior research shows that both factors influence model behavior, sometimes leading to substantial differences in performance \cite{durmus2024persuasion,nisbett_how_2023}. Our study provides a lower bound on LLMs' persuasive capability, as different prompts or larger models could yield alternative—and potentially stronger—persuasive outcomes. Future research should examine how variations in both prompt design and model size affect not only persuasion success but also user trust, engagement, and resistance, particularly in real-world, long-term interactions.

By addressing these limitations, future studies can refine our understanding of LLM-driven persuasion and its implications for human decision-making in everyday contexts.

\section{Acknowledgments}
We gratefully acknowledge the support of the Hebrew University Starting Grant and ISF grant No. 977/24.
We also thank Noam Siegelman and Asael Sklar for their invaluable insights, as well as Daria Lioubashevski for her generous assistance.

\bibliographystyle{apacite}

\setlength{\bibleftmargin}{.125in}
\setlength{\bibindent}{-\bibleftmargin}

\bibliography{general,culture,Sycophancy}

\end{document}